# Shape Interaction Matrix Revisited and Robustified: Efficient Subspace Clustering with Corrupted and Incomplete Data

Pan Ji[1], Mathieu Salzmann[2,3], and Hongdong Li[1,4]
[1]Australian National University, Canberra
[2]CVLab, EPFL, Switzerland; [3]NICTA, Canberra
[4]ARC Centre of Excellence for Robotic Vision (ACRV)

## Abstract

*The Shape Interaction Matrix (SIM) is one of the earliest approaches to performing subspace clustering (i.e., separating points drawn from a union of subspaces). In this paper, we revisit the SIM and reveal its connections to several recent subspace clustering methods. Our analysis lets us derive a simple, yet effective algorithm to robustify the SIM and make it applicable to realistic scenarios where the data is corrupted by noise. We justify our method by intuitive examples and the matrix perturbation theory. We then show how this approach can be extended to handle missing data, thus yielding an efficient and general subspace clustering algorithm. We demonstrate the benefits of our approach over state-of-the-art subspace clustering methods on several challenging motion segmentation and face clustering problems, where the data includes corrupted and missing measurements.*

## 1. Introduction

In this paper, we tackle the problem of subspace clustering, which consists of finding the subspace memberships of points drawn from a union of subspaces. This problem has attracted a lot of attention in the community due to its applicability to many different tasks, such as motion segmentation and face clustering.

Most of the research in this area takes its roots in the pioneering work of Costeira and Kanade [5], which introduced the Shape Interaction Matrix (SIM) to solve the motion segmentation problem, i.e., the problem of clustering point trajectories into the motions of multiple rigid objects. More specifically, the SIM was defined as the orthogonal projection matrix onto the row space of the trajectory matrix, and was proven to directly encode the motion membership of each trajectory. This result was later shown to extend to the general problem of subspace clustering [16, 17].

While the SIM provably yields perfect clusters given ideal measurements from independent subspaces, the quality of the clusters quickly degrades in the presence of noise, as illustrated by Fig. 1. As a consequence, many algorithms have been proposed to improve the robustness of subspace clustering. However, these methods typically work either by using discriminant criteria to reduce the effects of noise [14, 34], which may be sensitive to the noise level, or by formulating subspace clustering as a regularized optimization problem [9, 10, 15, 20, 21], thus requiring to tune the regularization weight to the data at hand. Furthermore, little work has been done to address the missing data scenario, for which, to the best of our knowledge, expensive two-steps methods (i.e., data completion followed by clustering) are typically employed [27, 32].

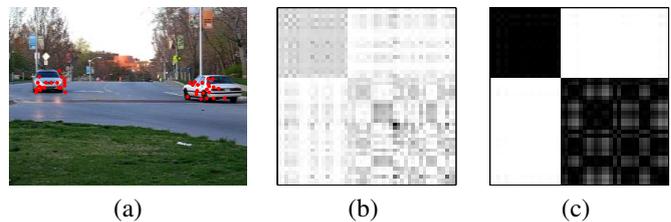

Figure 1: **Subspace clustering example:** (a) Two motions, each forming one subspace; (b) Shape Interaction Matrix of the trajectories in (a), which is sensitive to noise; (c) Affinity matrix obtained by our method: a much clearer block-diagonal structure.

In this paper, we revisit the use of the SIM for subspace clustering and study its connections to several recent algorithms. Based on our analysis, we show that simple, yet effective modifications of the SIM can significantly improve its robustness to data corruptions. This, in turn, lets us introduce an efficient approach to handling missing data, whose presence is inevitable in real-world scenarios.

We demonstrate the effectiveness of our algorithms on motion segmentation and face clustering in different scenarios, including the presence of noise, outliers and missing data. Our experiments evidence the benefits of our approach

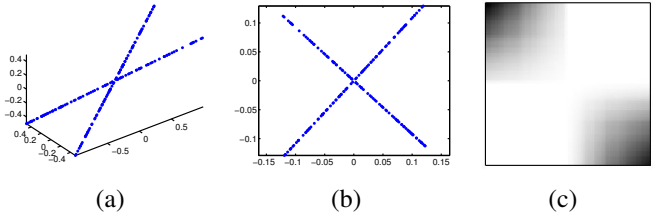

(a)  (b)  (c)

Figure 2: **SIM for clustering two lines in 3D: (a)** Two lines (each forming one subspace) with an arbitrary angle; **(b)** New data representation with $\mathbf{V}_r$. Note that the lines have become orthogonal; **(c)** SIM (absolute value) normalized by its maximum value; the darker the SIM image, the greater the value.

over existing methods in all these scenarios.

## 2. SIM Revisited: Review and Analysis

The Shape Interaction Matrix (SIM) was originally introduced by Costeira and Kanade [5] to extend Tomasi and Kanade's groundbreaking work [29] on factorization-based structure-from-motion from a single motion to the multibody case. In the single-motion scenario, the trajectory matrix $\mathbf{X} \in \mathbb{R}^{2F \times N}$ (for $N$ points in $F$ frames) can be factorized into the product of a motion matrix $\mathbf{M} \in \mathbb{R}^{2F \times 4}$ and a shape matrix $\mathbf{S} \in \mathbb{R}^{4 \times N}$ with metric (rotation and translation) constraints. However, for multi-body motions, the metric constraints no longer directly apply, but require the knowledge of the membership of each point to each motion.

In their work [5], Costeira and Kanade showed that these motion memberships could be obtained from the data itself. To this end, they introduced the SIM, defined as

$$\mathbf{Q} = \mathbf{V}_r \mathbf{V}_r^T \;, \quad (1)$$

where $\mathbf{V}_r \in \mathbb{R}^{N \times r}$ is the matrix containing the first $r$ right singular vectors of $\mathbf{X}$, with $r = 4K$ in the case of $K$ non-degenerate motions. Mathematically, the SIM is the *orthogonal projection matrix* onto the *column space* of $\mathbf{V}_r$, or, equivalently, onto the *row space* of $\mathbf{X}$. Importantly, it can be shown that $\mathbf{Q}_{ij} = 0$ if points $i$ and $j$ belong to different motions, and $\mathbf{Q}_{ij} \neq 0$ if points $i$ and $j$ belong to the same motion. Therefore, in [5], segmentation was achieved by block-diagonalizing $\mathbf{Q}$, which at the time involved an expensive operation.

Intuitively, we can think of $\mathbf{V}_r$ as a new data representation of the original $\mathbf{X}$, with each row of $\mathbf{V}_r$ a data point. Then, the theory of the SIM shows that different independent subspaces become *orthogonal* to each other in the new representation. In Fig. 2, we demonstrate this via a toy example.

The main drawback of the SIM arises from the fact that, while it yields provably correct clusters for independent motions and noise-free measurements, its accuracy decreases in the presence of noise, outliers, or degenerate motions. Over the years, many methods have therefore been proposed to improve the SIM. In the remainder of this section, we review these methods in a rough chronological order.

### 2.1. The Pre-Spectral-Clustering Era

Earlier approaches to accounting for noise, outliers and degeneracies [6, 11, 12, 14, 16, 17, 34, 36] were mostly focused on modifications of the SIM itself, or on directly related formulations. For instance, Gear [12] advocated the use of the reduced row echelon method instead of the SVD to better account for noise and automatically find the rank of the trajectory matrix. Wu et al. [34] presented an orthogonal subspace decomposition method to make the SIM more robust to noise by reasoning at group-level instead of considering individual point trajectories.

From a more general perspective, Kanatani [16, 17] reformulated motion segmentation as a subspace separation problem, and showed that under the condition that the subspaces are linearly independent, the SIM is block-diagonal (up to a permutation of the data). Later, Zelnik-Manor and Irani [36] considered the degenerate cases of the motion segmentation problem when the motions are not independent. They analyzed the causes of these degeneracies and proposed to overcome some of them by using the eigenvectors $\mathbf{E} = [\mathbf{e}_1^T \cdots \mathbf{e}_N^T]^T$ of the row-normalized matrix $\mathbf{X}^T \mathbf{X}$ and constructing a new shape interaction matrix as $\mathbf{Q}_{ij} = \sum_{k=1}^r \exp((\mathbf{e}_i(k) - \mathbf{e}_j(k))^2)$.

### 2.2. The Post-Spectral-Clustering Era

An important advance in the subspace clustering research was achieved by Park et al. [24], who, based on the then recent success of spectral clustering methods [23, 28], showed that the absolute value of the SIM could be employed as an affinity matrix in spectral clustering, thus yielding more accurate results than much more sophisticated methods, such as [18]. This then moved the focus of the subspace clustering community away from the SIM (at least in appearance, as discussed below) and towards designing better affinity matrices for spectral clustering.

In this context, Yan and Pollefeys [35] introduced a Local Subspace Affinity (LSA) measure to build affinity matrices. LSA measures the affinity between two points as the principal angle between their local subspaces. Instead of using the original data points $\mathbf{X}$, LSA represents the data with the row-normalized singular vectors $\mathbf{V}$ of $\mathbf{X}$. More recently, Lauer and Schnörr [19] proposed a spectral-clustering-based method that directly relies on the angles between the data points. As in LSA, instead of computing the angles from the original data, they also represented the data with its normalized singular vectors.

The recent trends in the subspace clustering literature exploit the notion of *self-expressiveness* of the data to build affinity matrices [9, 10, 15, 20, 21]. The idea of self-expressiveness was introduced in [9] to describe the fact that each data point can be represented as a linear combination of the other points. To exploit this idea to construct an affinity matrix, one has to ensure that such a linear combination for a point has non-zero coefficients only for the points in the same subspace. In other words, with the coefficients grouped in a matrix $\mathbf{C}$, $\mathbf{C}_{ij} = 0$ if points $i$ and $j$ belong to different subspaces, and $\mathbf{C}_{ij} \neq 0$ otherwise. This can be achieved by minimizing certain norms of $\mathbf{C}$. In particular, Sparse Subspace Clustering (SSC) [9, 10] considered the $\ell_1$ norm of $\mathbf{C}$; Low Rank Representation (LRR) [20, 21] the nuclear norm of $\mathbf{C}$; and Efficient Dense Subspace Clustering (EDSC) [15] the Frobenius norm of $\mathbf{C}$. Interestingly, in [15], it was shown that LRR and EDSC are equivalent to the SIM in the noise-free case. The difference lies in their ability to handle noise and outliers via additional regularization terms in their objective functions. Note that, even in the noisy case, it was shown [25] that the optimal solutions of LRR and EDSC take the form $\mathbf{V}\mathcal{P}(\Sigma)\mathbf{V}^T$, where $\mathcal{P}(\cdot)$ denotes the shrinkage-thresholding operator. Therefore, these solutions essentially correspond to a modified version of the SIM. More importantly, the effect of the regularizers introduced by these methods is sensitive to their weights, which therefore need to be tuned for the data at hand.

While many methods address the problem of robustness to noise with complete data, little work has been done to handle the missing data scenario with only a few exceptions such as [27, 32, 37, 38]. However, these methods [27, 32, 38] typically follow an expensive two-step procedure, i.e., data matrix completion followed by subspace clustering. In [37], the missing entries are simply set to zero so that they have no contribution in computing affinities by data correlations. This method, although directly handling missing data, does not make full use of the data itself because some observed entries are also discarded by the simple zeroing out strategy.

By contrast, we introduce an efficient subspace clustering method that is directly motivated by the SIM, but does not require additional regularization terms to handle data corruptions. More specifically, we show how the SIM can be robustified to data corruptions via three simple steps. We then further introduce an algorithm that robustly recovers the row-space of the data from incomplete measurements via an efficient iterative update on the Grassmann manifold, thus effectively making the powerful SIM representation applicable to the missing data scenario.

## 3. SIM Robustified: Corrupted Data

In this section, we introduce a robust subspace clustering method inspired by the SIM, but that lets us handle cor-rupted measurements.

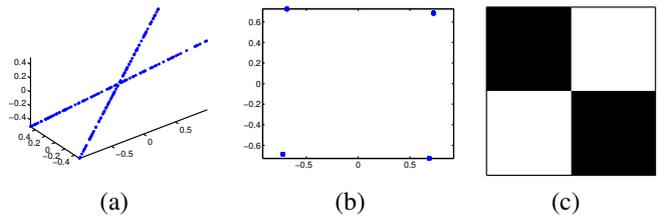

(a) (b) (c)

Figure 3: **Clustering two lines in 3D: Row normalization** (a) Two lines (each forming one subspace) with an arbitrary angle; (b) New data representation with row-normalized $\mathbf{V}_r$. Note that the lines collapse four points on the unit circle, corresponding to orthogonal vectors; (c) New SIM (absolute value) without magnitude bias.

To make the SIM method robust to data corruptions, we design a series of three steps: (i) row normalization of $\mathbf{V}_r$, (ii) elementwise powering of the new SIM, and (iii) determining the best rank $r$. While the first two steps aim at making direct modifications to the SIM, the third step is designed to account for degenerate cases, e.g., planar motions. In the remainder of this section, we present these steps and explain the rationale behind them.

### 3.1. Row Normalization

A closer look at the SIM method reveals that there is a *magnitude bias* within it, i.e., although the inter-cluster (subspace) affinities are guaranteed to be zero, the intra-cluster (subspace) affinities depend on the magnitude of data points. More specifically, for points drawn from the same subspace, the affinities between those that are closer to the origin will be smaller than between those that are further away. For example, in Fig. 2(c), the affinity values are much smaller in the center (i.e., points close to the origin) than in the corners. However, *ideally all points on the same subspace should be treated equally, since they belong to the same class*. Moreover, this magnitude bias is also undesirable because it makes the points close to the origin more sensitive to noise.

To avoid the magnitude bias, we introduce an extra step, row normalization of $\mathbf{V}_r$, so that all data points in the new representation have the same magnitude 1. As a consequence, in an ideal scenario, the new SIM will become uniform within each subspace, as illustrated in Fig. 3.

### 3.2. Elementwise Powering

In an ideal scenario (i.e., without noise), after row normalization the inter-cluster affinities are all zero and the intra-cluster affinities all one. However, in noisy cases, the elements of the affinity matrix (i.e., the absolute value of the new SIM) lie in the interval $[0, 1]$, and the inter-cluster affinities are often nonzero, but have rather small values. *Elementwise powering of the new SIM will thus virtually suppress these small values while keeping the large affinities mostly unaffected.* This operation is quite intuitive and just aims to denoise the SIM (affinity matrix). It was first used in [19] to increase the gap between intercluster and intra-cluster affinities. The result of this step is illustrated in Fig. 4. Since, after row normalization, the data in $\mathbf{V}_r$ always has a similar magnitude, independently of the problem of interest, the same powering factor can always be employed, thus preventing the need to tune a parameter for the data at hand. In our experiments, we observed that values in $[3, 5]$ generally yield good results.

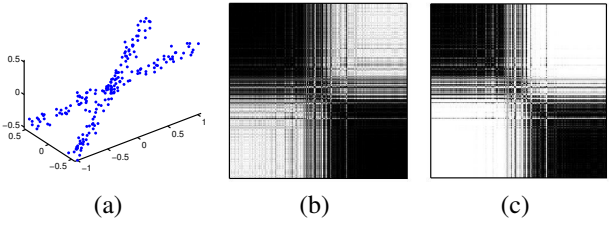

Figure 4: **Clustering two lines with noise in 3D: (a)** Two lines with Gaussian noise; **(b)** New SIM after row normalization, with noise in the off-diagonal blocks; **(c)** Affinity matrix after elementwise powering. Note that the block-diagonal structure is much cleaner.

**Remark** *The first two steps are simple and intuitive. However, we can also interpret them in a more theoretical way with kernel methods. Note that each row of $\mathbf{V}_r$ is a point in the new data representation, so the SIM $\mathbf{Q} = \mathbf{V}_r \mathbf{V}_r^T$ indeed consists of the inner product of every pair of these points. Now we show that the first two steps, i.e. row normalization and elementwise powering, are equivalent to applying a normalized polynomial kernel. Given a polynomial kernel $\kappa(\mathbf{v}_i, \mathbf{v}_j) = (\mathbf{v}_i^T \mathbf{v}_j)^\alpha$ and its corresponding feature mapping $\phi$, the normalized polynomial kernel $\hat{\kappa}(\mathbf{v}_i, \mathbf{v}_j)$ corresponds to the feature map*

$$\mathbf{v}_i \longrightarrow \frac{\phi(\mathbf{v}_i)}{\|\phi(\mathbf{v}_i)\|} \ . \tag{2}$$

*Then the kernel $\hat{\kappa}$ can be expressed as*

$$\hat{\kappa}(\mathbf{v}_i, \mathbf{v}_j) = \frac{\phi(\mathbf{v}_i^T)}{\|\phi(\mathbf{v}_i)\|} \frac{\phi(\mathbf{v}_j)}{\|\phi(\mathbf{v}_j)\|} \tag{3}$$

$$= \frac{\kappa(\mathbf{v}_i, \mathbf{v}_j)}{\sqrt{\kappa(\mathbf{v}_i, \mathbf{v}_i)\kappa(\mathbf{v}_j, \mathbf{v}_j)}} \tag{4}$$

$$= \frac{(\mathbf{v}_i^T \mathbf{v}_j)^\alpha}{\sqrt{(\mathbf{v}_i^T \mathbf{v}_i)^\alpha (\mathbf{v}_j^T \mathbf{v}_j)^\alpha}} \tag{5}$$

$$= \left( \frac{\mathbf{v}_i^T \mathbf{v}_j}{\sqrt{(\mathbf{v}_i^T \mathbf{v}_i)(\mathbf{v}_j^T \mathbf{v}_j)}} \right)^\alpha \tag{6}$$

$$= \left( \frac{\mathbf{v}_i^T}{\|\mathbf{v}_i\|} \frac{\mathbf{v}_j}{\|\mathbf{v}_j\|} \right)^\alpha , \tag{7}$$

*where the last equality indicates two steps, i.e. normalization and powering.*

### 3.3. Rank Determination

Determining the correct rank $r$ is crucial for the success of the SIM method. As early as in [12], it was shown that the SIM yields poor results if the incorrect rank is employed. Therefore, several approaches to determining the correct rank have been studied. In [6] the rank was obtained by examining the gaps in the singular values, which is typically sensitive to the level of noise. Inspired by the sparse representation community, ALC [22] uses the sparsity-preserving dimension $d_{\rm sp} = \min d$ s.t. $d \geq 2D \log(2F/d)$, where $D$ is the estimated intrinsic dimensionality of each subspace. SC [19] estimates the rank by looking at the relative eigenvalue gaps of the Laplacian matrix.

Here, we draw inspiration from the matrix perturbation theory and introduce a simple, yet effective method to detect the correct rank of the SIM. In general, one can easily define a range of possible ranks $[r_{\min}, r_{\max}]$. Our rank selection method then works by simply exhaustively searching over all possible rank values, and selecting the $r$ which minimizes

$$C(r) = \frac{\text{minCut}(A_1^r, \cdots, A_K^r)}{|\lambda_K - \lambda_{K+1}|} \ , \tag{8}$$

where $A_i^r$ is the $i^{\text{th}}$ cluster of the graph defined by the affinity matrix $\mathbf{A}^r$, $\lambda_i$ is the $i^{\text{th}}$ largest eigenvalue of the Laplacian matrix $\mathbf{L}^r = \mathbf{D}^{-1} \mathbf{A}^r$ (where $\mathbf{D}$ is the degree matrix of $\mathbf{A}^r$), and the minimal cut $\text{minCut}(A_1^r, \cdots, A_K^r)$ can be obtained via the Ncuts algorithm [28]. Intuitively, the smaller the $\text{minCut}$ and the larger the eigengap, the better the segmentation.

Our rank selection criterion can be justified by the Davis-Kahan Theorem from the matrix perturbation theory, which provides an upper bound on the distance between the eigenspaces of two Hermitian matrices that differ by some perturbations. This theorem is stated below.

**Theorem 1 (Davis-Kahan Theorem [7])** *Let $\mathbf{L}$ and $\widetilde{\mathbf{L}}$ be two N-by-N Hermitian matrices. Let $\{\lambda_1, \cdots, \lambda_k, \lambda_{k+1}, \cdots, \lambda_N\}$ ($\lambda_i \geq \lambda_j, i < j$) denote the eigenvalues of $\mathbf{L}$, and $\mathbf{U}_1$ the matrix containing its first $k$ eigenvectors. Let $\{\tilde{\lambda}_1, \cdots, \tilde{\lambda}_k, \tilde{\lambda}_{k+1}, \cdots, \tilde{\lambda}_N\}$ and $\tilde{\mathbf{U}}_1$ be the analogous quantities for $\widetilde{\mathbf{L}}$. Then, by defining $\sigma := \min_{1 \leq i \leq k, 1 \leq j \leq n-k} |\lambda_i - \tilde{\lambda}_{k+j}|$, we have*

$$\| \sin \Theta(\mathbf{U}_1, \tilde{\mathbf{U}}_1) \|_F \leq \frac{\|\widetilde{\mathbf{L}} - \mathbf{L}\|_F}{\sigma} \ , \tag{9}$$

*where $\Theta(\mathbf{U}_1, \tilde{\mathbf{U}}_1)$ is the vector of principal angles between $\mathbf{U}_1$ and $\tilde{\mathbf{U}}_1$.*

**Algorithm 1** Robust Shape Interaction Matrix (RSIM)

---

**Input:** Data matrix $\mathbf{X}$, minimum rank $r_{\min}$, and maximum rank $r_{\max}$

**for** $r := r_{\min}$ **to** $r_{\max}$ **do**

1. *SVD*: Compute the SVD of the data matrix $\mathbf{X}$, i.e., $\mathbf{X} = \mathbf{U}\mathbf{\Sigma}\mathbf{V}^T$, and take the first $r$ right singular vectors $\mathbf{V}_r$.
2. *Normalization*: Normalize each row of $\mathbf{V}_r$ to have unit norm $\to \widetilde{\mathbf{V}}_r$.
3. *New SIM*: Build the new Shape Interaction Matrix as $\mathbf{Q} = \widetilde{\mathbf{V}}_r \widetilde{\mathbf{V}}_r^T$.
4. *Powering*: Take the elementwise power of $\mathbf{Q}$, i.e., $\mathbf{A}_{ij} = (\mathbf{Q}_{ij})^\gamma$.
5. *Rank Determination*: Apply the normalized cuts algorithm to get the cluster labels, and compute the value $C(r)$ as in Eq. 8.

**end for**

$r_{\text{best}} = \underset{r}{\operatorname{argmin}} C(r)$.

**Output:** The cluster labels $\mathbf{s}$, the best rank $r_{\text{best}}$.

---

The Davis-Kahan Theorem states that the distance between the eigenspaces of two Hermitian matrices that differ by some perturbations is bounded by the ratio between the perturbation level and their eigengap. In our case, since we do not have access to the true Laplacian, we make use of the eigenvalues of the noisy Laplacian to estimate the eigengap $\sigma$, which will then occur between the $K^{th}$ and $K + 1^{th}$ eigenvalues for $K$ clusters. Furthermore, we rely on minCut to approximate the noise level of the Laplacian matrix $\mathbf{L}$. This approximation is reasonable because $\mathbf{L}$ is nothing but a normalized version of the affinity matrix. So by minimizing $C(r)$, we aim to find the lowest upper bound of the distance between the noisy Laplacian and the true one. This minimum should correspond to the optimal rank.

### 3.4. Robust Shape Interaction Matrix

Our complete Robust Shape Interaction Matrix (RSIM) algorithm is outlined in Algorithm 1. Note that, while its steps are simple, to the best of our knowledge, it is the first time that such an algorithm is proposed. Furthermore, our experiments clearly evidence the effectiveness of RSIM and its benefits over more sophisticated methods, such as SSC and LRR.

## 4. SIM Robustified: Missing Data

Our previous solution to handling data corruption relies on the computation of the row space $\mathbf{V}$ of the data $\mathbf{X}$. When the data contains missing entries, computing the row-space cannot simply be achieved by SVD. Here, we exploit the idea that our goal truly is to estimate the subspace on which the data lies (which $\mathbf{V}$ is an orthogonal basis of). Linear subspaces of a fixed rank form a Riemannian manifold known as the Grassmannian. Therefore, we propose to make use of an optimization technique on the Grassmann manifold to obtain an estimate of $\mathbf{V}$ in the presence of missing data.

More formally, let $\mathcal{G}(N, r)$ denote the Grassmann manifold of $r$-dimensional linear subspaces of $\mathbb{R}^N$ [4][1]. A point $\mathbf{Y} \in \mathcal{G}(N, r)$, i.e., an $r$-dimensional subspace of $\mathbb{R}^N$, can be represented by any orthogonal matrix $\mathbf{V} \in \mathbb{R}^{N \times r}$ whose columns span the $r$-dimensional subspace $\mathbf{Y}$. Estimating the row space $\mathbf{V}$ (an orthogonal matrix) of the data matrix can then be thought of as finding the corresponding linear subspace on $\mathcal{G}(N, r)$.

To estimate $\mathbf{V}$, we utilize the GROUSE (Grassmannian Rank-One Update Subspace Estimation) algorithm [1]. GROUSE is an efficient online algorithm that recovers the column space of a highly incomplete observation matrix. To this end, it utilizes a gradient descent method on the Grassmannian to incrementally update the subspace by considering one column of the observation matrix at a time.

More specifically, in our context, at each iteration $t$, we take as input a vector $\mathbf{x}_{\Omega_t} \in \mathbb{R}^{N_t}$, which corresponds to the partial observation of a single vector $\mathbf{x}_t \in \mathbb{R}^N$ in the data matrix $\mathbf{X}$[2], with observed indices defined by $\Omega_t \subset \{1, \cdots, N\}$. Let $\mathbf{V}_{\Omega_t}$ be the submatrix of $\mathbf{V}$ consisting of the rows indexed by $\Omega_t$. Following the GROUSE formalism, which relies on the least-squares reconstruction of the data, we can formulate the update at iteration $t$ as the solution to the optimization problem

$$\min_{\mathbf{a}, \mathbf{V}} \frac{1}{2} \|\mathbf{V}_{\Omega_t} \mathbf{a} - \mathbf{x}_{\Omega_t}\|_2^2 \tag{10}$$
$$\text{s. t.} \quad \mathbf{V}^T \mathbf{V} = \mathbf{I}_{r \times r} \, ,$$

where $\mathbf{a}$ corresponds to the representation (or weights) of the data $\mathbf{x}_{\Omega_t}$ in the current estimate of the subspace, and $\mathbf{I}_{r \times r}$ is the identity matrix.

Since (10) is not jointly convex in $\mathbf{a}$ and $\mathbf{V}$, the two variables are obtained in a sequential manner: First, the optimal weights $\mathbf{w}$ are computed for the current subspace, and then the subspace is updated given those weights. Due to the least-squares form of the objective function, the solution for the weights can be obtained in closed-form as $\mathbf{w} = \mathbf{V}^{\dagger}_{\Omega_t} \mathbf{x}_{\Omega_t}$, where $\mathbf{V}^{\dagger}_{\Omega_t}$ is the pseudoinverse of $\mathbf{V}_{\Omega_t}$. To update the subspace, i.e., the orthogonal basis matrix $\mathbf{V}$, GROUSE exploits an incremental gradient descent method on the Grassmann manifold, which we describe below.

Let $\mathbf{I}_{\Omega_t} \in \mathbb{R}^{N \times N_t}$ be the $N_t$ columns of the $N \times N$ identity matrix indexed by $\Omega_t$. Then, the objective function

---

[1] For example, the Grassmann manifold $\mathcal{G}(N, 1)$ consists of all lines in $\mathbb{R}^N$ passing through the origin.

[2] Note that even though we consider $\mathbf{x}_t$ to be a column vector, it really corresponds to one row of the data matrix $\mathbf{X}$.

of (10) can be rewritten as

$$E_t = \|\mathbf{I}_{\Omega_t}(\mathbf{V}_{\Omega_t}\mathbf{w} - \mathbf{x}_{\Omega_t})\|_2^2 \ . \quad (11)$$

The update of the subspace is achieved by taking a step in the direction of the gradient of this objective function on the Grassmannian, i.e., moving along the geodesic defined by the negative Grassmannian gradient. To this end, we first need to compute the regular gradient of the objective function with respect to $\mathbf{V}$. This gradient can be written as

$$\frac{\partial E_t}{\partial \mathbf{V}} = -(\mathbf{I}_{\Omega_t}(\mathbf{x}_{\Omega_t} - {}_{\Omega_t}\mathbf{w}))\mathbf{w}^T \quad (12)$$

$$= -\mathbf{r}\mathbf{w}^T \ , \quad (13)$$

where $\mathbf{r} = \mathbf{I}_{\Omega_t}(\mathbf{x}_{\Omega_t} - \mathbf{V}_{\Omega_t}\mathbf{w})$ denotes the (zero-padded) vector of residuals.

The gradient on the Grassmannian can then be obtained by projecting the regular gradient on the tangent space of the Grassmannian at the current point. Following [1,8], this can be written as

$$\nabla E_t = (\mathbf{I} - \mathbf{V}\mathbf{V}^T)\frac{\partial E_t}{\partial \mathbf{V}} \quad (14)$$

$$= -(\mathbf{I} - \mathbf{V}\mathbf{V}^T)\mathbf{r}\mathbf{w}^T \quad (15)$$

$$= -\mathbf{r}\mathbf{w}^T \ . \quad (16)$$

As shown in [8], a gradient step along the geodesic with tangent vector $-\nabla E_t$ is defined as a function of the singular values and vectors of $\nabla E_t$. Since $\nabla E_t$ has rank one, its singular value decomposition is trivial to compute. This lets us write a step of length $\eta$ in the direction $-\nabla E_t$, and thus the update of $\mathbf{V}$ at time $t$, as

$$\mathbf{V}_{t+1} = \mathbf{V}_t + \frac{(\cos(\sigma\eta) - 1)}{\|\mathbf{w}\|^2}\mathbf{V}\mathbf{w}\mathbf{w}^T + \sin(\sigma\eta)\frac{\mathbf{r}}{\|\mathbf{r}\|}\frac{\mathbf{w}^T}{\|\mathbf{w}\|} \ , \quad (17)$$

where $\sigma = \|\mathbf{r}\|\|\mathbf{w}\|$.

The Grassmannian update is very efficient since each subspace update only involves linear operations. Furthermore, for a specific diminishing step-size $\eta$, it is guaranteed to converge to a locally optimal estimate of $\mathbf{V}$ [1]. After getting an estimate of $\mathbf{V}$ using this method, we can directly apply the RSIM to perform subspace clustering.

The pseudocode of our robust SIM with missing data (RSIM-M) algorithm is given in Algorithm 2. Note that:

1. Stochastic gradient descent may require a relatively large number of steps to be stable. With small amounts of data, we run multiple passes over the data. For example, in our experiments on motion segmentation with incomplete trajectories, we iterated over all the frames 100 times. Thanks to the high efficiency of rank-one Grassmannian update, RSIM-M remains very efficient.

---

**Algorithm 2** RSIM with Missing Data (RSIM-M)

**Input:** An incomplete data matrix $\mathbf{X}$, a subspace initialization $\mathbf{V}_0$, a step size $\eta$, bounds $r_{\min}, r_{\max}$

   **for** t = 1,$\cdots$,T **do**
     1. Take the $t^{\text{th}}$ row of $\mathbf{X}$ with observed entry $\Omega_t$.
     2. Update the current $\mathbf{V}_t$ via Eq. 17.
   **end for**
   Run Algorithm 1 to perform robust subspace clustering.

**Output:** The cluster labels $\mathbf{s}$, the best rank $r_{\text{best}}$.

---

2. Due to the non-convexity of this problem, initialization is important for convergence speed and optimality. In practice, we start with the subspace spanned by the most complete $r$ rows of $\mathbf{X}$, which we found to be very effective in practice.

## 5. Experimental Evaluation

We evaluate the performance of our algorithms with four sets of experiments that represent different scenarios: (i) Hopkins155 for motion segmentation; (ii) Extended Yale Face B for face clustering; (iii) Hopkins12Real: 12 additional real-world sequences with missing data; (iv) Hopkins outdoor sequences for semi-dense motion segmentation. We compare the results of our algorithms with the following baselines: SIM (followed by spectral clustering) ( [24]), SSC ( [10]), LRSC ( [31]), LRR ( [20]), and EDSC [15]. Note that the last two methods have proposed to make use of an additional post-processing step (called a heuristic in [10]), which yields the additional baselines LRR-H and EDSC-H. For the case of LRR, this heuristic relies on the following steps:

1. Solve the optimization problem

$$\min_{\mathbf{C},\mathbf{E}} \|\mathbf{C}\|_* + \lambda\|\mathbf{E}\|_{2,1} \quad \text{s.t.} \quad \mathbf{X} = \mathbf{XC} + \mathbf{E} \ . \quad (18)$$

2. Compute the SVD of $\mathbf{C}$, i.e. $\mathbf{C} = \mathbf{U}\Sigma\mathbf{V}^T$, and take the first $r$ singular vectors $\mathbf{V}_r$.
3. Construct $\mathbf{Z} = \mathbf{V}_r \Sigma_r^{\frac{1}{2}}$, and normalize each row of $\mathbf{Z}$.
4. Build the affinity matrix $\mathbf{A}$ as $\mathbf{Z}\mathbf{Z}^T$ with elementwise powering such that $\mathbf{A}_{ij} = [\mathbf{Z}\mathbf{Z}^T]_{ij}^4$.

Interestingly, this post-processing is nothing else but another way to build an improved SIM. Indeed, $\mathbf{Z}$ can be thought of approximately as the row space of the denoised data $\mathbf{X} - \mathbf{E}$ from the equality constraint in (18). In other words, one can also think of LRR (and EDSC) as a pre-processing step to denoise the data before computing the SIM. In contrast, the proposed method does not require any pre-processing step and, as evidenced below, achieves much better results.

The parameters of the baselines are tuned to the best results for each experiment. For our method, we report the results of all the four sets of experiments with the same powering factor $\gamma = 4.5$. Note that we could potentially get better results if we fine-tuned the parameter $\gamma$. For motion segmentation, the rank is selected iteratively from the integers in $[K, 4K]$; for the face clustering experiment, the rank is in $[4K, 6K]$ with $K$ the number of clusters.

### 5.1. Hopkins155: Complete Data with Noise

Hopkins155 [30] is a standard benchmark to test point-based motion segmentation algorithms. It includes 155 sequences, each of which contains 39-550 point trajectories sampled from two or three motions. Each trajectory is complete and contaminated with a moderate amount of noise, but with no outliers. The dataset contains general motions, such as rigid and nonrigid motions, indoor checkerboard sequences and outdoor traffic sequences. The results of our RSIM algorithm and of the baselines are reported in Table 1. Note that our method achieves the lowest overall average clustering error. The average runtimes (in seconds) per sequence for different methods are: SIM – 0.0229s, SSC – 0.9187s, LRR – 1.0795s, LRR-H – 1.0930s, EDSC – 0.0378s, EDSC-H – 0.0762s, and RSIM – 0.1766s.

We also performed an ablation study on this dataset to see the contributions of the proposed steps. We denote the SIM with our first two steps (i.e., normalization and polynomial kernel) by SIM+1&2, and denote the SIM with our third step (i.e., rank determination) by SIM+3. The results are shown in Table 2. Note that the proposed first two steps improve the motion segmentation accuracy over the original SIM, the proposed third step boosts the segmentation results with a big margin, and our complete robust shape interaction matrix method achieves the best results.

### 5.2. Extended Yale B: Complete Data with Outliers

Under Lambertian reflectance assumption, face images of the same subject with a fixed pose and varying lighting lie approximately in a low dimensional subspace [2]. We therefore make use of the Extended Yale B face dataset to evaluate our method on the task of face clustering. This dataset is composed of face images of 38 subjects, each of which has 64 frontal face images acquired under different lighting conditions. We follow exactly the same experimental settings as in [10] and divide the 38 subjects into four groups (i.e., group 1 - subject 1 to 10, group 2 - subject 11 to 20, group 3 - subject 21 to 30, and group 4 - subject 31 to 38). Within each group, we test all the combinations of $K$ subjects, for $K \in [2, 3, 5, 8, 10]$.

Note that, since this data is grossly corrupted, the baselines ([10, 15, 20]) use an additional regularizer to account for outliers, with weight specifically tuned for this dataset. In contrast, our method doesn't have this extra term and parameter. The results are presented in Table 3. Interestingly, although our method does not handle the outliers explicitly, it achieves the comparable accuracies for 2 and 3 subjects, and get far better accuracies for 5, 8 and 10 subjects. In contrast to the baselines, our method remains stable as the number of subjects increases. From a different perspective, this dataset can be thought of as being contaminated with both Gaussian noise and Laplacian noise, so the baseline methods (SSC, LRR and EDSC) all have two regularization terms, one for the Gaussian noise and the other for the Laplacian one, and their weight parameters, therefore, need to be tuned for the data at hand. In contrast, our method relies on no specific assumptions about the distributions of the noise, and is thus robust to a mixture of different types of noise.

### 5.3. Hopkins12Real: Incomplete Data with Noise

To demonstrate that our method can handle missing data gracefully, we employed the Hopkins 12 additional sequences containing incomplete data and noise. Most of the baselines used previously cannot deal with missing data. Therefore, we only compare our method with those that have proposed to tackle this challenging scenario. In particular, we compare our results against those published by [26], where ALC was employed after filling in the missing entries of the data matrix with a matrix completion method, e.g., Power Factorization (PF) ([13]), Robust Principal Component Analysis (RPCA) ([3]), and $\ell_1$ sparse representation ([26]). We also evaluate SSC ([9, 10]), which works with missing data by either removing the trajectories with missing entries (SSC-R), or treating the missing entries as outliers (SSC-O). In contrast, our method doesn't require any matrix completion or trajectory removal. The results in Table 4 clearly evidence the benefits of our method in the presence of missing data.

### 5.4. Hopkins Outdoor: Semi-dense, Incomplete Data with Outliers

To study a more realistic scenario, where outliers and missing data are ubiquitous due to occlusions and tracking failures, we took 18 outdoor sequences from the Hopkins155 dataset and obtained semi-dense trajectories by applying the tracking method of [33][3]. For the 18 sequences, the tracking method found an average of 3026 trajectories per sequence, among which 16.66% (684 out of 3026) on average contained missing entries, which were set to zero. We compare our results to those of the same SSC-O and SSC-R baselines used previously.

Since there is no ground-truth for this data, we can only provide a qualitative comparison. In particular, we observed that our method performed either better, or on par with SSC-R, and consistently outperformed SSC-O. We found that SSC-O tends to group the trajectories with missing entries in a single cluster. This is mainly due to the fact that, according to the self-expressiveness criterion, incomplete trajectories are poorly represented by complete ones, and thus end up being grouped together. Fig. 5 shows some typical behaviors of SSC-R and of our approach. It can easily be checked that our approach yields better clusters on average. The results of SSC-O are shown in Fig. 6, where the behavior described above can be observed. Finally, in Fig. 7, we show some failure cases where both SSC-R and our approach were unable to find the right clusters. The results for all the sequences are provided in the appendix. Since SSC-R removes the missing trajectories, it utilized only 2522 trajectories on average out of the original average of 3026. In contrast, our method makes use of all the available trajectories. Nonetheless, while SSC-R takes 150.48 seconds per sequence on average, our method only takes about 5.22 seconds on average.

---
[3]While there are 21 outdoor videos in Hopkins155, the tracking code that we used was unable to read the 3 Kanatani videos.

Table 1: Clustering error (in %) on Hopkins 155.

| Methods | SIM | SSC | LRR | LRR-H | EDSC | EDSC-H | **RSIM** |
|---|---|---|---|---|---|---|---|
| **2 motions** | | | | | | | |
| Mean | 6.50 | 1.53 | 4.10 | 2.13 | 2.67 | 0.86 | **0.65** |
| Median | 1.14 | **0.00** | 0.22 | **0.00** | **0.00** | **0.00** | **0.00** |
| **3 motions** | | | | | | | |
| Mean | 12.26 | 4.40 | 9.89 | 4.03 | 8.06 | 2.49 | **1.71** |
| Median | 6.12 | 6.22 | 0.56 | 1.43 | 2.53 | **0.21** | 0.28 |
| **Overall** | | | | | | | |
| Mean | 7.80 | 2.18 | 5.41 | 2.56 | 4.04 | 1.23 | **0.89** |
| Median | 1.53 | **0.00** | 0.53 | **0.00** | 0.30 | **0.00** | **0.00** |

Table 2: Ablation study on Hopkins 155.

| Methods | SIM | SIM+1&2 | SIM+3 | **RSIM** |
|---|---|---|---|---|
| Mean | 7.80 | 5.77 | 3.17 | **0.89** |
| Median | 1.53 | 0.24 | 0.31 | **0.00** |

## 6. Conclusion

In this paper, we have revealed that many recent subspace clustering methods actually did not go far beyond the 20-year-old SIM method, but rather had indirect connections to it. While recent methods exploit notions of compressed sensing and self-expressiveness, our method performs simple and direct modifications of the SIM itself and makes it robust to corruptions. Furthermore, we have extended our method to the case of missing data. Our experimental evaluation has demonstrated that our algorithms are not only efficient, but also generally applicable to subspace segmentation in realistic scenarios. In the future, we plan to adapt our method to online motion segmentation on longer sequences.

## Acknowledgements

NICTA is funded by the Australian Government through the Department of Communications and the Australian Research Council through the ICT Centre of Excellence Program. HL thanks the supports of ARC Discovery grants DP120103896, DP130104567, and the ARC Centre of Excellence.

Table 3: Clustering error (in %) on Extended Yale B.

| Methods | SIM | SSC | LRR | LRR-H | EDSC | EDSC-H | **RSIM** |
|---|---|---|---|---|---|---|---|
| **2 subjects** | | | | | | | |
| Mean | 8.10 | **1.86** | 9.52 | 2.54 | 5.42 | 2.65 | 2.36 |
| Median | 6.25 | **0.00** | 5.47 | 0.78 | 4.69 | 1.56 | 1.56 |
| **3 subjects** | | | | | | | |
| Mean | 24.64 | **3.10** | 19.52 | 4.21 | 14.05 | 3.86 | 3.21 |
| Median | 16.67 | **1.04** | 14.58 | 2.60 | 8.33 | 3.13 | 2.60 |
| **5 subjects** | | | | | | | |
| Mean | 45.62 | 4.31 | 34.16 | 6.90 | 36.99 | 5.11 | **3.56** |
| Median | 48.13 | **2.50** | 35.00 | 5.63 | 30.63 | 3.75 | 3.13 |
| **8 subjects** | | | | | | | |
| Mean | 57.05 | 5.85 | 41.19 | 14.34 | 54.24 | 6.07 | **3.60** |
| Median | 55.96 | 4.49 | 43.75 | 14.34 | 48.73 | 4.88 | **3.32** |
| **10 subjects** | | | | | | | |
| Mean | 65.10 | 10.94 | 38.85 | 22.92 | 59.58 | 7.24 | **3.70** |
| Median | 64.06 | 5.63 | 41.09 | 23.59 | 50.47 | 6.09 | **3.44** |

Table 4: Clustering error (in %) on Hopkins 12 Real Motion Sequences with Incomplete Data.

| % | PF+ALC | RPCA+ALC | $\ell_1$+ALC | SSC-R | SSC-O | **RSIM-M** |
|---|---|---|---|---|---|---|
| Mean | 10.81 | 13.78 | 1.28 | 3.82 | 8.78 | **0.61** |
| Median | 7.85 | 8.27 | 1.07 | **0.31** | 4.80 | 0.61 |
| Max | 34.57 | 41.36 | 4.35 | 20.25 | 26.34 | **1.64** |
| Std | 0.04 | 12.25 | 1.29 | 6.80 | 8.79 | **0.53** |

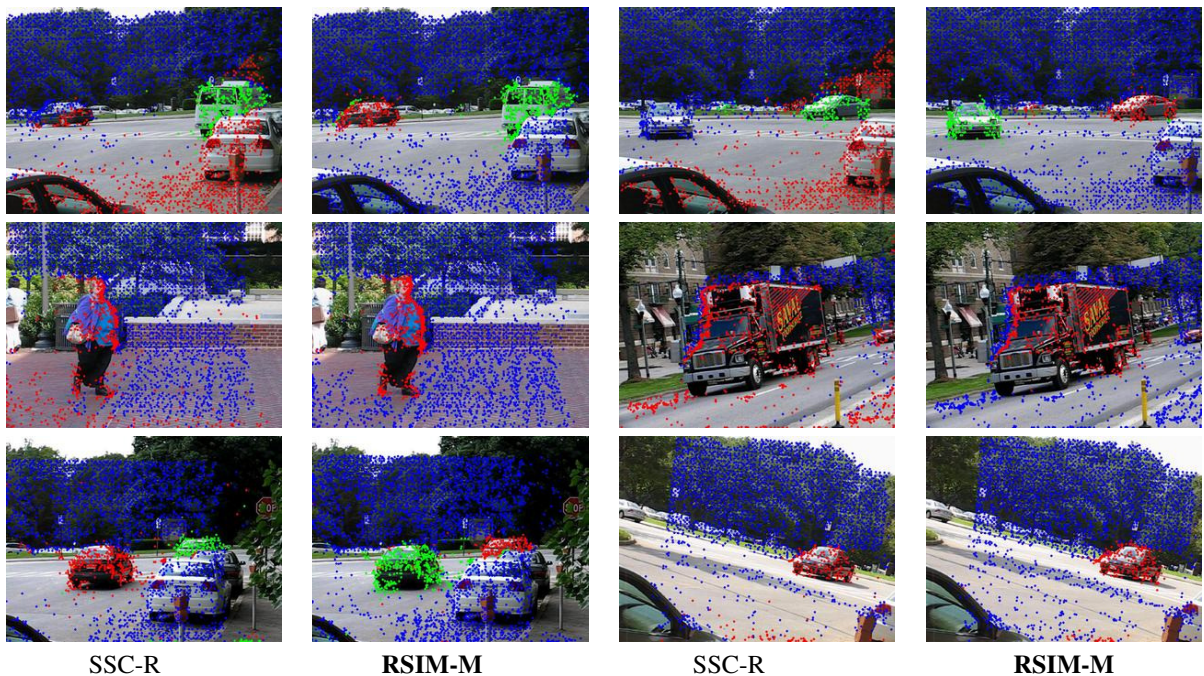

| SSC-R | **RSIM-M** | SSC-R | **RSIM-M** |

Figure 5: **Comparison of SSC-R and RSIM-M on semi-dense data:** While SSC-R removes the trajectories with missing entries, and thus gets less dense results, our method can handle missing data robustly. Each image is a frame sampled from one of the video sequences. The points marked with the same color are clustered into the same group by the respective methods. Best viewed in color.

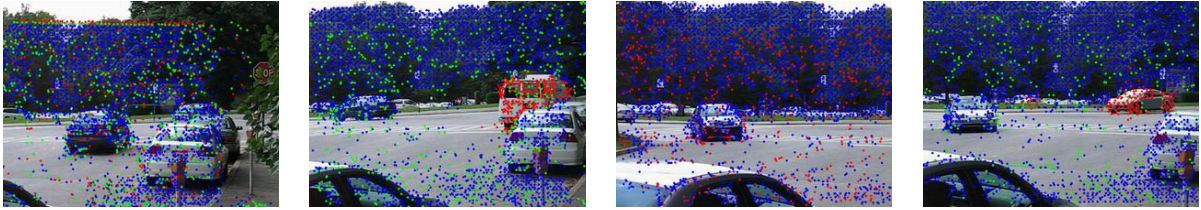

Figure 6: **Typical behavior of SSC-O on semi-dense data:** By treating missing entries as outliers, SSC-O tends to cluster the trajectories with missing entries into same group. The points marked with the same color are clustered into the same group by SSC-O. Best viewed in color.

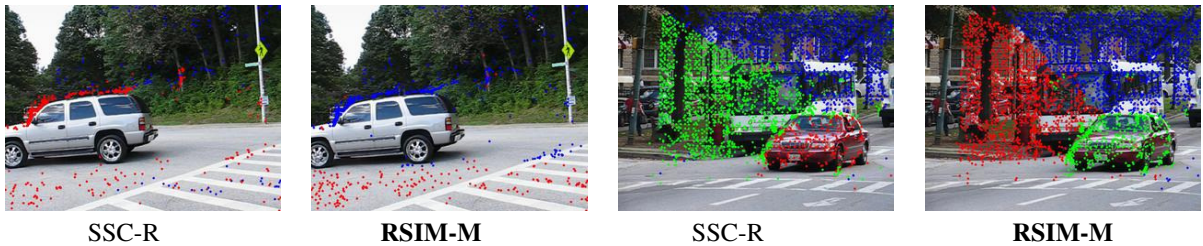

SSC-R　　　　**RSIM-M**　　　　SSC-R　　　　**RSIM-M**

Figure 7: **Failure cases of SSC-R and of RSIM-M**: We conjecture that these failures are due to tracking failures (e.g., very few trajectories), or to highly dependence between motions. Best viewed in color.

# Appendix – Hopkins Outdoor: Semi-dense, Incomplete Data with Outliers

Here, we show the results of our algorithm (RSIM-M) and of the baselines SSC-O and SSC-R on all the 18 sequences described in Section 5.4 of this paper.

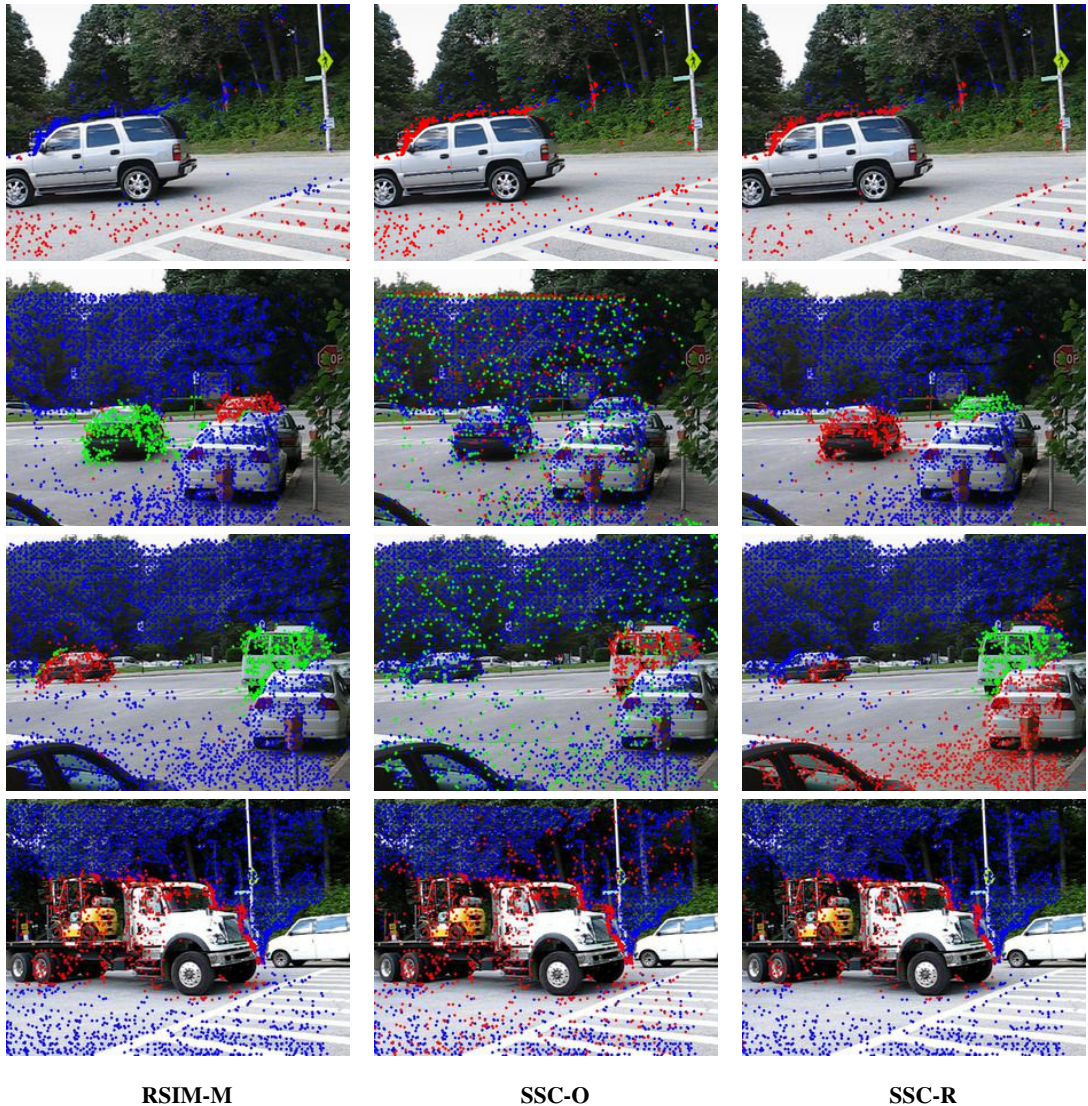

|      RSIM-M      |      SSC-O      |      SSC-R      |

Figure 8: Semi-dense motion segmentation results for sequences 1-4. Our method (RSIM-M) uses all the available tracks (**3026 on average**) with an **average runtime of 5.22 seconds per sequence**; SSC-O tends to group the trajectories with missing entries in the same cluster; SSC-R takes 150.48 seconds on average and only makes use of 2522 points on average after removing the incomplete trajectories. Best viewed in color.

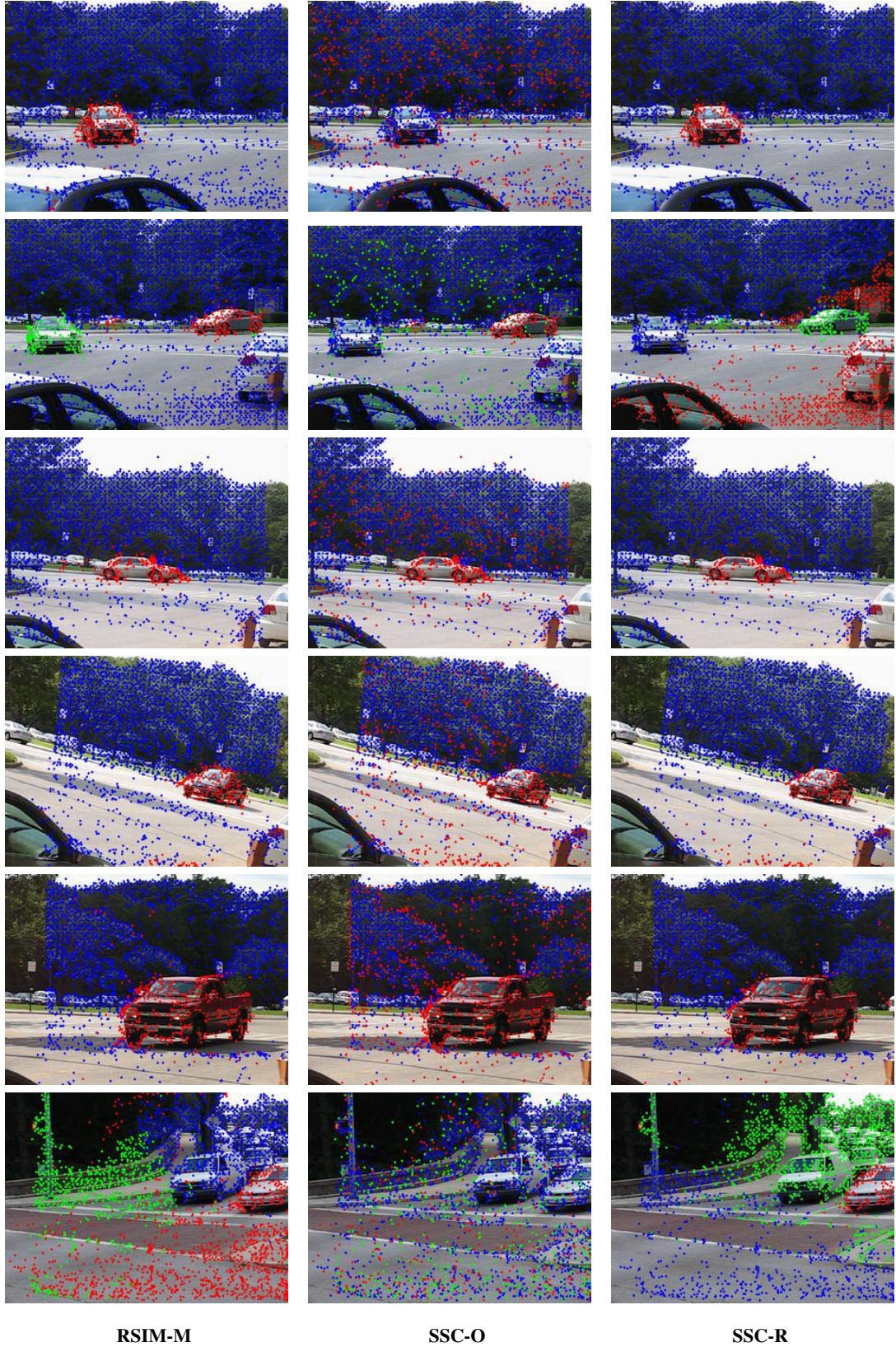

**RSIM-M**         **SSC-O**         **SSC-R**

Figure 9: Semi-dense motion segmentation results for sequences 5-10. Our method (RSIM-M) uses all the available tracks (**3026 on average**) with an **average runtime of 5.22 seconds per sequence**; SSC-O tends to group the trajectories with missing entries in the same cluster; SSC-R takes 150.48 seconds on average and only makes use of 2522 points on average after removing the incomplete trajectories. Best viewed in color.

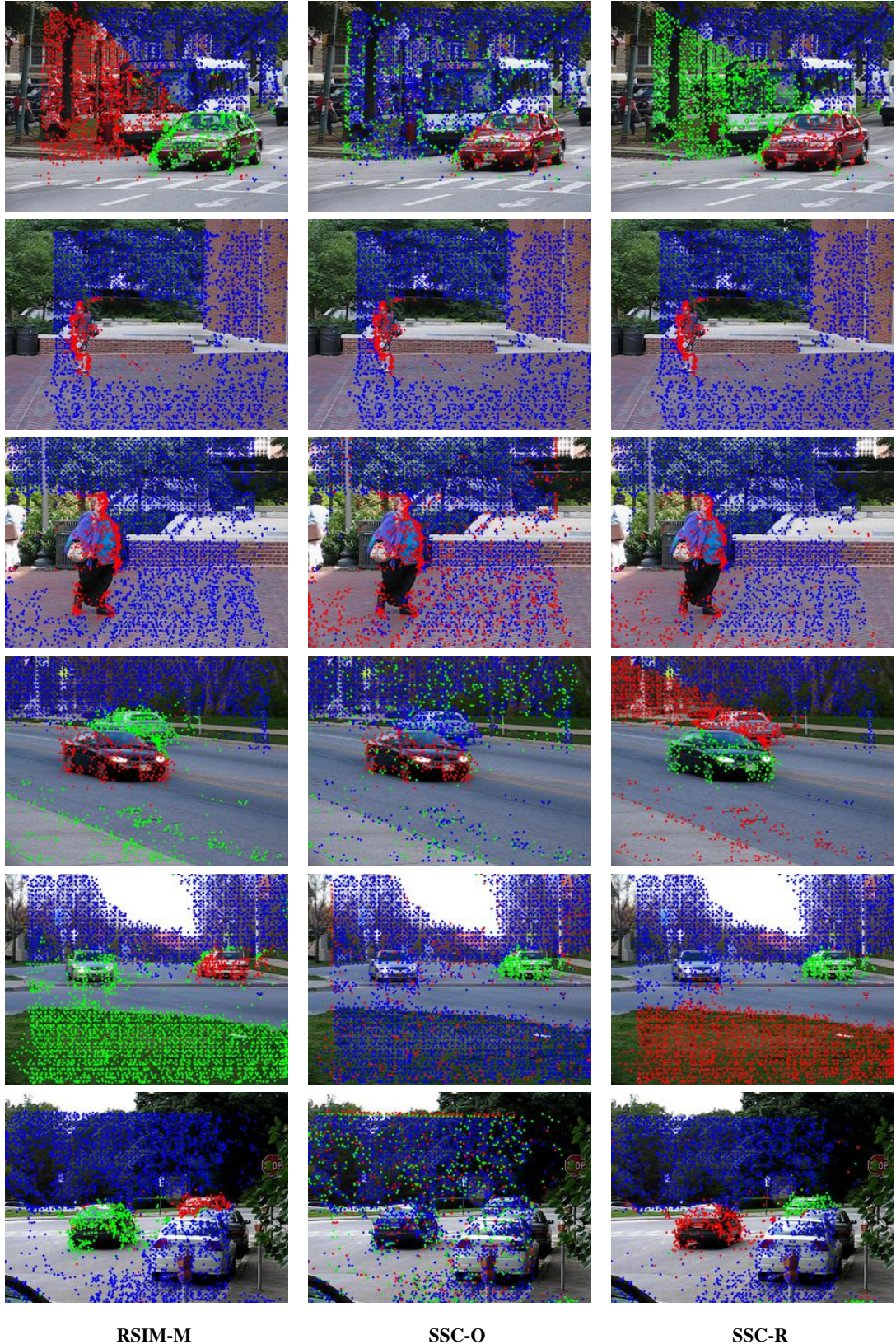

Figure 10: Semi-dense motion segmentation results for sequences 11-16. Our method (RSIM-M) uses all the available tracks (**3026 on average**) with an **average runtime of 5.22 seconds per sequence**; SSC-O tends to group the trajectories with missing entries in the same cluster; SSC-R takes 150.48 seconds on average and only makes use of 2522 points on average after removing the incomplete trajectories. Best viewed in color.

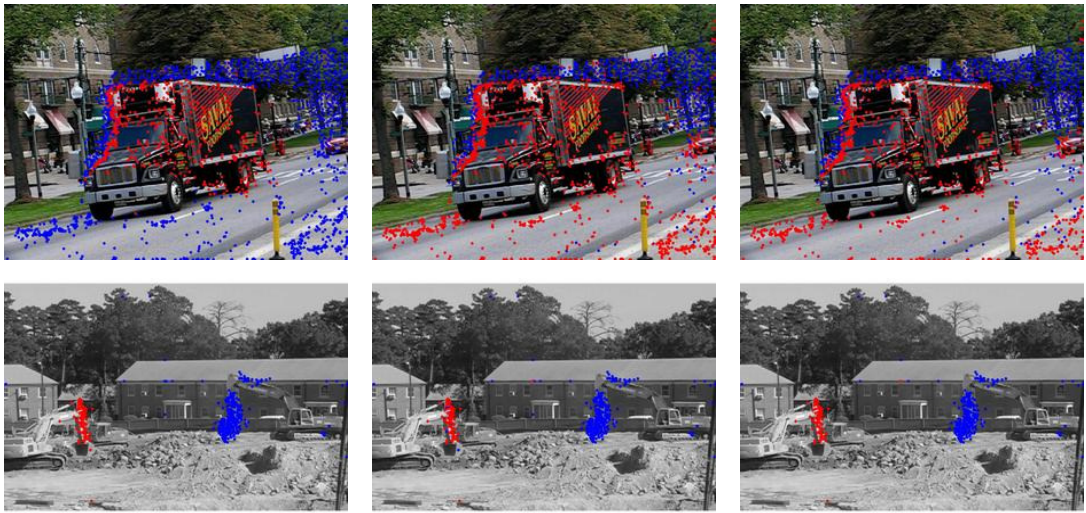

**RSIM-M**          **SSC-O**          **SSC-R**

Figure 11: Semi-dense motion segmentation results for sequences 17-18. Our method (RSIM-M) uses all the available tracks (**3026 on average**) with an **average runtime of 5.22 seconds per sequence**; SSC-O tends to group the trajectories with missing entries in the same cluster; SSC-R takes 150.48 seconds on average and only makes use of 2522 points on average after removing the incomplete trajectories. Best viewed in color.